\documentclass[times, review, 10pt]{elsarticle}

\usepackage{amssymb}
\usepackage{amsmath}
\usepackage{algorithmic}
\usepackage{algorithm}
\usepackage{array}
\usepackage[caption=false,font=normalsize,labelfont=sf,textfont=sf]{subfig}
\usepackage{textcomp}
\usepackage{stfloats}
\usepackage{url}
\usepackage{verbatim}
\usepackage{graphicx}
\usepackage{natbib}
\usepackage{multirow}
\usepackage{tabularx}
\usepackage{booktabs}
\usepackage{threeparttable}
\usepackage[table]{xcolor}
\definecolor{grey}{RGB}{230,230,230}

\journal{Nuclear Physics B}

\begin{document}

\begin{frontmatter}



\title{EPIR: An Efficient Patch Tokenization, Integration and Representation Framework for Micro-expression Recognition}

\author[label1]{Junbo Wang\fnref{1}}
\ead{jbwang@nwpu.edu.cn}
\author[label1]{Liangyu Fu\fnref{1}}
\ead{lyfu@mail.nwpu.edu.cn}
\author[label1]{Yuke Li\corref{*}}
\ead{liyuke@nwpu.edu.cn}
\affiliation[label1]{organization={School of Software, Northwestern Polytechnical University},
            city={Xi'an},
            postcode={710129},
            country={China}}
\author[label2]{Yining Zhu}
\ead{yiningzhu@nwpu.edu.cn}
\affiliation[label2]{organization={School of Computer Science, Northwestern Polytechnical University},
            city={Xi'an},
            postcode={710129},
            country={China}}
\author[label3]{Xuecheng Wu}
\ead{wuxc3@stu.xjtu.edu.cn}
\affiliation[label3]{organization={School of Computer Science and Technology, Xi'an Jiaotong University},
            city={Xi'an},
            postcode={710049},
            country={China}}
\author[label4]{Kun Hu}
\ead{hukun\_sdu@hotmail.com}
\affiliation[label4]{organization={School of Science, Edith Cowan University},
            city={WA},
            postcode={6027},
            country={Australia}}

\cortext[*]{Corresponding author}
\fntext[1]{Both authors contributed equally to this work.}



\begin{abstract}
Micro-expression recognition can obtain the real emotion of the individual at the current moment. Although deep learning-based methods, especially Transformer-based methods, have achieved impressive results, these methods have high computational complexity due to the large number of tokens in the multi-head self-attention. In addition, the existing micro-expression datasets are small-scale, which makes it difficult for Transformer-based models to learn effective micro-expression representations. Therefore, we propose a novel Efficient Patch tokenization, Integration and Representation framework (EPIR), which can balance high recognition performance and low computational complexity. Specifically, we first propose a dual norm shifted tokenization (DNSPT) module to learn the spatial relationship between neighboring pixels in the face region, which is implemented by a refined spatial transformation and dual norm projection. Then, we propose a token integration module to integrate partial tokens among multiple cascaded Transformer blocks, thereby reducing the number of tokens without information loss. Furthermore, we design a discriminative token extractor, which first improves the attention in the Transformer block to reduce the unnecessary focus of the attention calculation on self-tokens, and uses the dynamic token selection module (DTSM) to select key tokens, thereby capturing more discriminative micro-expression representations. We conduct extensive experiments on four popular public datasets (i.e., CASME~II, SAMM, SMIC, and CAS(ME)$^3$). The experimental results show that our method achieves significant performance gains over the state-of-the-art methods, such as 9.6\% improvement on the CAS(ME)$^3$ dataset in terms of UF1 and 4.58\% improvement on the SMIC dataset in terms of UAR metric.
\end{abstract}




\end{frontmatter}



\section{Introduction}
Humans intuitively convey psychological states through facial expressions, which involve the activation of various facial muscle regions to communicate emotions. These expressions can be classified into macro-expressions and micro-expressions based on the extent of muscle movement~\cite{paul2007emotions}. Micro-expression recognition, a specialized task within affective computing, focuses on classifying brief micro-expression clips into distinct emotional categories. In contrast to macro-expressions, micro-expressions are characterized by their spontaneous nature, subtle intensity, and rapid duration, typically lasting between 1/25 to 1/3 of a second~\cite{li2022deep}.

\begin{figure}[t]
\centering
\includegraphics[width=0.9\textwidth]{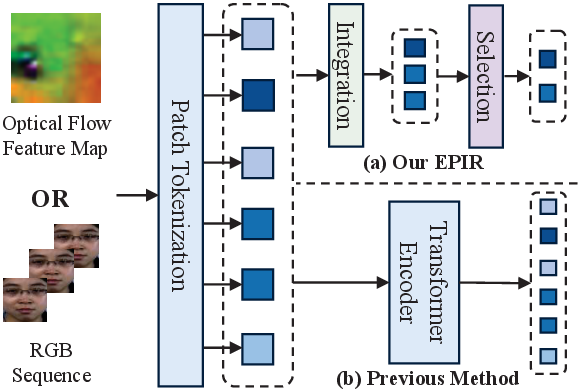}
\caption{Comparison of (a) our EPIR and (b) the previous Transformer-based micro-expression recognition model. The blue square denotes a visual token.}
\label{fig:intro}
\end{figure}

As a result, micro-expressions serve as reliable indicators of genuine emotions, even when individuals attempt to suppress or mask their true feelings. This unique characteristic makes micro-expression recognition particularly valuable across various critical domains, including psychological evaluation, commercial negotiations, and human-computer interaction~\cite{li2022deep}. However, the accurate recognition presents significant challenges due to their inherent characteristics: involuntary nature, extremely rapid duration, and subtle spatiotemporal variations in facial muscle movements. Furthermore, micro-expression recognition can be influenced by contextual emotional cues and culturally specific expression patterns, adding additional complexity to recognition~\cite{merghani2018review},~\cite{crivelli2019inside},~\cite{niedenthal2019historical}.

Recently, various feature representation learning methods, such as hand-crafted features with expert experiences and deep learning-based methods~\cite{pfister2011recognising, wang2014lbp, li2020joint, wei2023cmnet} have been explored. However, these methods, especially Transformer-based methods, suffer from high computational complexity due to excessive tokens in the self/cross-attention mechanism, as shown in Fig.~\ref{fig:intro} (b). At the same time, since micro-expression recognition is a fine-grained task, the main purpose is to extract a few key tokens that can represent a certain micro-expression class. Excessive unimportant tokens will bring noise and affect the model's deterministic judgment. In addition, the current performance ceiling of micro-expression recognition methods is constrained by the limited scale of available datasets, primarily due to the inherent challenges in acquiring authentic spontaneous micro-expression samples under controlled experimental conditions, which consequently makes it challenging for conventional transformer architectures with substantial data requirements to learn effective representations for micro-expression recognition. 

To address these challenges, we propose a novel Efficient Patch tokenization, Integration and Representation framework (EPIR) with Vision Transformer for micro-expression recognition. 
It is constructed with a dual norm shifted patch tokenization (DNSPT) module, a token integration module, and a subsequent discriminative token extractor composed of a dynamic token selection module (DTSM) and cascading Transformer block with an inter-token self-attention, as illustrated in Fig.~\ref{fig:overall_diagram}. To effectively process each sample, our EPIR extracts optical flow features using a pair of frames - the onset and apex frames from the input micro-expression clip. Unlike previous methods that directly use convolutional neural networks (CNNs) or Transformers for representation, our EPIR first utilizes the DNSPT module to learn spatial relations between neighboring facial pixels through spatial transformation and dual norm projection. 
Next, it uses the token integration module to fuse tokens with similar information, reducing the number of tokens involved in the calculation while ensuring recognition performance. In order to extract key tokens that can represent micro-expression classes from numerous token sequences, we designed a discriminative token extractor, that uses the cascading Transformer block to represent micro-expression information, incorporating DTSM to select the key patches/tokens, and in each transformer block, EPIR improved the self-attention to achieve more deterministic decisions. 
Notably, the overall EPIR is trained end-to-end with a subsequent MLP head and can dynamically generate more discriminative representations, even with small-scale micro-expression datasets. The experimental results show that our method achieves significant performance gains over the state-of-the-art methods.

It should be noted that this paper is an extended version of our conference paper PTSR~\cite{fu2025ptsr} presented at ACM ICMR 2025. The journal version includes the new design to the method for micro-expression recognition, we want to explore more possibilities for model efficiency. Thus, we propose the token integration in the Transformer blocks to further lightweight the model while guaranteeing recognition performance; To fully verify the recognition performance of the model, we conducted more experiments. In addition to the traditional experiments under the Composite Database Evaluation (CDE) protocol, we carried out a larger scope of experiments under the Single Database Experiments (SDE) protocol on each dataset, and carried out visual qualitative analysis, so as to evaluate the model performance more comprehensively and objectively.

In summary, the overall contributions of this paper are summarized as follows:
\begin{itemize}
    \item We propose a dual norm shifted patch tokenization module to learn spatial relation between neighboring pixels of the face region, which is implemented by elaborating spatial transformation and dual norm projection.
    \item We propose a token integration module to reduce the number of tokens in attention calculation while ensuring recognition performance.
    \item We design a discriminative token extractor composed of inter-token attention-based Transformer blocks and a dynamic token selection module (DTSM) to flexibly extract micro-expression key tokens.
    \item Extensive experiments on CASME II, SAMM, SMIC and CAS(ME)$^3$ datasets demonstrate the superiority of our EPIR, exhibiting the highest improvements of 4.58\% in terms of UAR and 3.5\% in terms of UF1 under the CDE protocol, 5.52\% in terms of UAR and 9.66\% in terms of UF1 under the SDE protocol.
\end{itemize}

\section{Related Work}
\label{sec:related}

\subsection{Traditional Micro-Expression Recognition}

Traditional micro-expression recognition relies on manually designed feature extraction and machine learning algorithms for classification. \cite{pfister2011recognising} employed temporal interpolation models and the first comprehensive spontaneous micro-expression corpus to accurately recognize. \cite{wang2014lbp} proposed a simple and effective Main Directional Mean Optical-flow (MDMO) feature for micro-expression recognition. \cite{liu2015main} proposed the Local Binary Patterns with Six Intersection Points (LBP-SIP) volumetric descriptor, which provided a more compact and lightweight representation. \cite{he2017multi} develop a novel multi-task mid-level feature learning method to enhance the discrimination ability of extracted low-level features by learning a set of class-specific feature mappings, which would be used for generating our mid-level feature representation. These methods basically follow the paradigm of face detection-feature extraction-feature selection-classification. They have achieved some success in simpler scenarios, but they often exhibit low robustness when faced with complex environments, such as variations in lighting, facial occlusions, or blurred expressions. Moreover, the limitations of hand-crafted feature design present challenges in handling the diversity and complexity of micro-expressions.

\subsection{Deep Learning-based Micro-Expression Recognition}

Deep learning supports end-to-end training and can directly learn from raw image or video data to the final micro-expression classification results, avoiding the information loss and error accumulation that may be caused by multiple independent steps such as feature extraction, feature selection, and classifier training in traditional methods. Recently, various deep learning-based methods have achieved groundbreaking advancements in micro-expression recognition, which utilize deep neural networks to automatically learn micro-expression representations, significantly enhancing recognition performance. From the perspective of deep neural network, these methods can be classified into CNN-based and Vision Transformer-based methods:

\textbf{CNN-based micro-expression recognition.} CNN has been extensively utilized in various computer vision downstream tasks since its proposal. Its robust local perception has significantly contributed to advancements in micro-expression recognition. CNN-based micro-expression recognition can be classified into three classes based on the data input modality: static images, dynamic sequence/features, and multi-modal combinations. 
The methods using static images typically select the apex frame of an micro-expression clip as the modeling target, allowing the model to learn the most typical micro-expression features, thus achieving high evaluation metrics \cite{li2020joint, li2018can}. However, these methods neglect the dynamic micro-expression information, making it unsuitable for certain applications, e.g., real-time emotion monitoring and human-computer interaction. 
Methods using dynamic sequence/features extract features from micro-expression image sequences and onset-apex pairs using 3D CNNs, or apply 2D CNNs to optical flow feature maps~\cite{zhou2019dual},~\cite{liong2019shallow},~\cite{xia2019spatiotemporal},~\cite{wei2023cmnet}. They focus on the essence of micro-expressions, thus holding greater research value from a practical application perspective. However, due to the ambiguity of dynamic features, the model must identify subtle discriminative features from numerous highly similar dynamic features for classification. As a result, the performance is often less than satisfactory. 
To enhance performance, some methods employ multi-stream CNN to extract both dynamic and static features, subsequently combining them using various strategies \cite{kumar2021micro, liu2020offset, sun2019two}. \cite{zhang2026multimodal} propose a novel multimodal learning framework that combines ME and physiological signals information. These methods compensate for the performance deficiencies of dynamic feature methods, but invariably have a significant increase in computational complexity.

\textbf{Vision Transformer-based micro-expression recognition.} Compared with CNNs, Vision Transformer \cite{dosovitskiy2020image} excels at modeling global features, which demonstrates better performance than CNNs in classification tasks. Recently, many studies have adopted Vison Transformers as the backbone for micro-expression recognition, achieving performance improvements~\cite{zhang2022short},~\cite{indolia2023micro} despite persistent issues caused by data modalities. However, due to the lack of local inductive bias in models compared to CNNs, the high performance of vision Transformers often comes at the cost of large-scale datasets. \cite{wang2024htnet} proposed HTNet to enhance recognition accuracy by segmenting micro-expressions into key areas. \cite{cai2024mfdan} proposed MFDAN to incorporate optical flow priors into the attention learning in the guided image encoding branch, enabling the model to focus on the most discriminative facial regions. Unfortunately, the limited scale of micro-expression datasets has been a long-standing challenge in micro-expression recognition due to the difficulty of collecting micro-expression samples. This leads to the frequent occurrence of overfitting during training in Transformer-based model for micro-expression recognition, ultimately hindering the improvement of recognition performance. Several works mitigated the issue through self-supervised learning \cite{nguyen2023micron} or by incorporating macro-expression data \cite{xia2020learning}. However, achieving accurate micro-expression recognition using supervised learning with fixed small-scale micro-expression datasets remains a critical study. At the same time, the high computational overhead caused by excessive tokens should also be addressed.

\section{Methodology}

\begin{figure}[t]
\begin{center}
   \includegraphics[width=\textwidth]{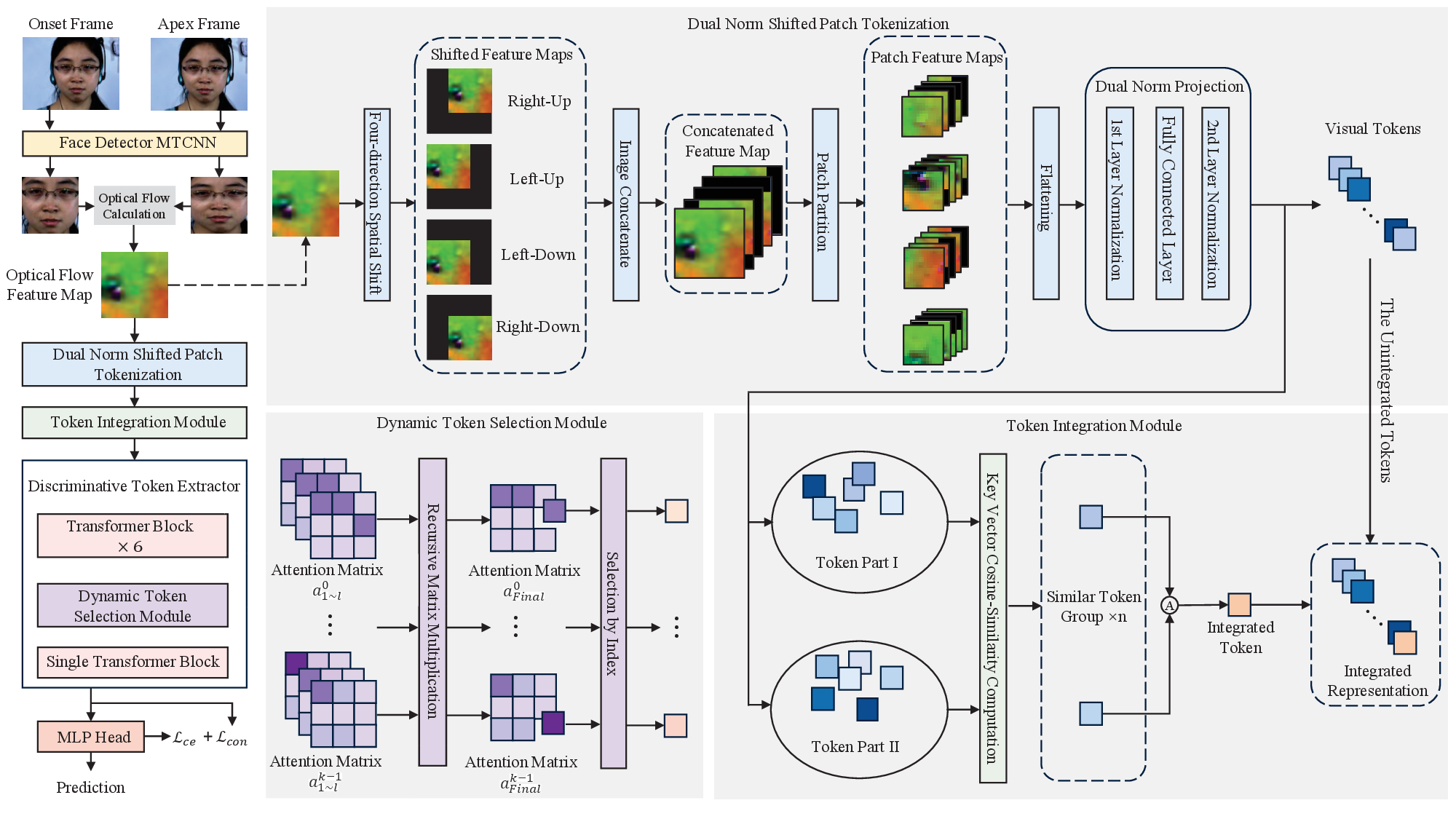}
\end{center}
   \caption{The overall framework of our proposed EPIR. The left column shows the overall framework of EPIR, and the right column shows the detailed working process of its three main modules. $\mathcal{L}_{ce}$ and $\mathcal{L}_{con}$ denote the cross-entropy and contrastive loss function, respectively.}
\label{fig:overall_diagram}
\end{figure}

Fig.~\ref{fig:overall_diagram} illustrates the overall framework of the proposed EPIR. First, we extract the onset frame \(X_{onset}\) (usually the first frame of the clip) and apex frame \(X_{apex}\) (the frame with the largest scale of facial muscle movement) from the micro-expression clips. Using MTCNN \cite{zhang2016joint}, we detect the facial regions from these two key frames, and we can obtain two RGB frames $X_{onset}^D$ and $X_{apex}^D$ that do not contain background noise. Next, we calculate the optical flow feature maps $V$ of the two RGB frames and input $V$ into DNSPT for path tokenization, thereby obtaining the initial representation of the micro-expression clips. Subsequently, we input the representation into the token integration module and the discriminative token extractor to get the discriminative tokens necessary for distinguishing different categories of micro-expressions. Finally, we input these discriminative tokens to the MLP Head for micro-expression recognition.

\subsection{Facial Optical Flow}

We use the Farneback's algorithm \cite{farneback2003two} to compute the optical flow between the two micro-expression frames as follows:
\begin{equation}
    V_{xy} = (u(x, y), v(x, y)),
\end{equation}

\begin{equation}
    V_z = \sqrt{\frac{{\partial V_x}^2}{\partial x} + \frac{{\partial V_y}^2}{\partial y} + \frac{1}{2} \left( \frac{{\partial V_x}^2}{\partial y} + \frac{{\partial V_y}^2}{\partial x} \right)}, 
\end{equation}
where $x \in (1, 2, ..., H)$ and $y \in (1, 2, ..., W)$ denote the height $H$ and width $W$ of the $X_i^D$, respectively, $u(x, y)$ and $v(x, y)$ are the horizontal and vertical components of the displacement estimate for each pixel in $X_i^D$, $V_{xy} = [V_x, V_y]$; $V_{xy} \in R^{W \times H \times 2}$, $\frac{{\partial V_x}^2}{\partial x}$, $\frac{{\partial V_y}^2}{\partial y}$, $\frac{{\partial V_x}^2}{\partial y}$, and $\frac{{\partial V_y}^2}{\partial x}$ are the first-order partial derivatives of $V_{xy}$. Finally, the optical flow feature map is formed as $V = [V_{xy}, V_z]$, $V \in R^{W \times H \times 3}$.

\subsection{Tokenization and Integration for Micro-Expression}

It is challenging to construct a large-scale micro-expression dataset due to the special nature of micro-expressions, so the current micro-expression datasets are generally small in scale, which affects the performance of deep learning-based micro-expression recognition. Therefore, how to make the model converge well and learn high-quality knowledge when trained from scratch on a small-scale dataset becomes a problem worth studying. To adapt EPIR to the small-scale datasets without using additional data, inspired by \cite{lee2021vision}, we propose Dual Norm Shifted Patch Tokenization (DNSPT).

\textbf{Dual Norm Shifted Patch Tokenization} makes the model learn spatial relation between neighboring pixels of the face region. As shown in Fig.~\ref{fig:overall_diagram}, first, the input optical flow feature map $V$ is spatially shifted by half of the feature map size in the four diagonal directions (i.e., right-up, left-up, left-down, and right-down). Next, the shifted feature maps are cropped to the same size as the input and concatenated to $V$. After that, the concatenated feature maps are divided into patches. Finally, the visual token $t$ is obtained by the dual norm projection module. The whole process is: 


\begin{equation}
    t = \mathrm{LN} \left( \mathrm{FC} \left( \mathrm{LN} \left( \mathrm{P} \left( \left[V;V_s^1;V_s^2;V_s^3;V_s^4 \right] \right) \right) \right) \right),
\end{equation}
where $V_s^i \in R^{H \times W \times 3}$ denotes the $i^{th}$ shifted optical flow feature map $V$, $\mathrm{P}(\cdot)$ denotes patch partition, $\mathrm{FC}(\cdot)$ denotes the learnable linear projection implemented by a fully connected layer, and $\mathrm{LN}(\cdot)$ denotes the layer normalization.

Next, we concatenate the class token with the visual token and then add the position embedding. The process can be expressed as:


\begin{equation}
    t_{pe} = [x_{cls}; t] + E_{pos},
\end{equation}
where $x_{cls} \in R^d$ denotes a class token, $E_{pos} \in R^{(N+1) \times d}$ denotes a learnable position embedding, $d$ is the hidden dimension of the Transformer block, and $N$ is the number of embedded tokens, $t_{pe} \in R^{(N+1) \times d}$.

\textbf{Token Integration Module.}
An excessive number of redundant tokens adds computational overhead without improving recognition performance.
Therefore, we proposed the token integration module. 

As shown in Fig.~\ref{fig:overall_diagram}, it consists of cascaded Transformer blocks, and we integrated tokens between the attention module and the feedforward neural network of each Transformer block. Set $t \in R^{N \times d}$ is the tokens output by a Transformer block in the token integration module, where $N$ is the number of tokens and $d$ is the dimension. We divide $t$ into two parts of equal capacity $t_1 \in R^{N/2 \times d}$ and $t_2 \in R^{N/2 \times d}$. Next, we take the similarity of the key vector of each token in the attention module as the token similarity, and calculate the similarity of all tokens in $t_1$ and $t_2$ pairwise. Based on the similarity, we select the most similar $K$ token pairs $t_{sim} \in R^{2 \cdot K \times d}$, and calculate the average of the two tokens in each pair to get the new integrated tokens $t_{int} \in R^{1 \times d}$, which will be concatenated with the unintegrated tokens $t_{res} \in R^{(N-K) \times d}$ and input into the feedforward neural network. The process can be expressed as:

\begin{equation}
    t_{sim}^{k} = [t_1^{i}; t_2^{j}] ,
\end{equation}
\begin{equation}
    t_{sim} = [t_{sim}^{1}; t_{sim}^{2}; ...; t_{sim}^{K}] ,
\end{equation}
\begin{equation}
    t_{res} = t[\sim t_{sim}] ,
\end{equation}
\begin{equation}
    t_{int} = \mathrm{AVG}(t_{sim}) ,
\end{equation}
\begin{equation}
    t = [t_{int}; t_{res}].
\end{equation}

\subsection{Discriminative Token Extraction}

Micro-expression differences often lie in the fine-grained pixel-level patterns, so the key to recognition is to extract fine-grained representations that effectively distinguish emotion categories. 
For Transformer-based models, this typically involves a small number of patches/tokens. Too many tokens may cause noise to the task and increase computational complexity. To extract key tokens for micro-expression recognition, we propose the discriminative token extractor. Specifically, it consists of a cascaded Transformer block and a dynamic token selection module. 
In addition, we improved the self-attention mechanism in the Transformer block by masking the self-token's attention score, thereby encouraging the model to focus on the relationships between the current token and other tokens. 
This module is described in detail below:

\textbf{Inter-Token Attention with Learnable Scaling (ITALS).} In the original Vision Transformer~\cite{dosovitskiy2020image}, the similarity matrix of multi-head self-attention can be expressed as:

\begin{equation}
    \label{eq:10}
    \mathrm{R_{i, j}}(x) = xE_q(xE_k)^T,
\end{equation}
where $x \in R^{(N+1) \times d}$ denotes the intermediate representation $\mathrm{LN}(Y_l)$ input to the multi-head self-attention, $E_q \in R^{d \times d_q}$ and $E_k \in R^{d \times d_k}$ denote the projection of the Query and Key, $d_q$ and $d_k$ is the dimensions of the Query and Key. $\mathrm{R_{i, j}}(x) \in R^{(N+1) \times (N+1)}$ denotes the similarity matrix calculated by the Query and Key.

In ITALS, we mask the diagonal elements, which assign higher inter-token relations scores by excluding self-token relations from participating in softmax operations. Specifically, the diagonal mask sets the diagonal component of the similarity matrix in the Query and Key computation to $-\infty$. The ITALS can more focus on other tokens than its own. The proposed diagonal mask is defined as follows:

\begin{equation}
    \mathrm{R_{i, j}^M}(x) =
\begin{cases}
\mathrm{R_{i, j}}(x) & (i \neq j)  \\
-\infty & (i = j)
\end{cases},
\end{equation}
where $\mathrm{R_{i, j}^M}(x)$ denotes each component of the masked similarity matrix.

The scaling in softmax controls the smoothness of the output distribution, the smaller the scaling, the sharper the distribution, which affects the diversity and certainty of the predictions. By ITALS, EPIR learns to adjust this scaling during training, improving the attention score by shrinking the scaling. This makes the output distribution more discriminative, thereby improving the accuracy and certainty of the predictions. The ITALS is defined as:

\begin{equation}
   \mathrm{ITALS}(x) = \mathrm{Softmax}(\frac{\mathrm{R^M_{i, j}}(x)}{\tau})xE_v,
\end{equation}
where $x$ is the same as $x \in R^{(N+1) \times d}$ in Eq.~\ref{eq:10}, $\tau$ is the learnable scaling, $E_v \in R^{d \times d_v}$ is the projection of the Value in self attention, $d_v$ is the dimension of the Value in self attention.

Set $t \in R^{(N-6 \cdot K) \times d}$ is the representation output by the token integration module. Next, we input $t$ into the 6 Transformer blocks, and the process can be expressed as:

\begin{align}
    Y_1^{'} &= \mathrm{ITALS}(\mathrm{LN}(t)), \\
    Y_1 &= \mathrm{FFN}(\mathrm{LN}(Y_1^{'})) + Y_1^{'}, \\
    Y_l^{'} &= \mathrm{ITALS}(\mathrm{LN}(Y_{l-1})) + Y_{l-1}, \label{eq:7} \\
    Y_l &= \mathrm{FFN}(\mathrm{LN}(Y_l^{'})) + Y_l^{'}, \label{eq:8}
\end{align}
where $Y_l^{'}$ denotes the intermediate representation in each Transformer block, $Y_l$ denotes the output representation of the $l^{th}$ Transformer block, and $l \in (2, 3, ..., 12)$. $\mathrm{FFN}(\cdot)$ denotes the feed-forward neural network, and $\mathrm{ITALS}(\cdot)$ denotes inter-token attention with learnable scaling.

\textbf{Dynamic Token Selection Module.} To address the ability of local discriminative representation capture, inspired by
\cite{he2022transfg}, we designed the dynamic token selection module (DTSM) before the last Transformer block in the discriminative token extractor. Fig.~\ref{fig:overall_diagram} shows how the attention matrix is integrated with DTSM. Specifically, set the number of ITALS heads to $K=3$, the intermediate representations input to the last Transformer block can be denoted as $Y_l=[Y^0_l; Y^1_l; Y^2_l; ...; Y^N_l]$, $l=12$. The attention weights of the 12 Transformer blocks before the last Transformer block are denoted as:

\begin{equation}
    A_l=[a^0_l; a^1_l; a^2_l; ...; a^K_l]\ l \in 1, 2, ..., L-1,
\end{equation}
\vspace{-2pt}
\begin{equation}
    a^i_l=[a^{i_0}_l; a^{i_1}_l; a^{i_2}_l; ...; a^{i_N}_l]\ i \in 0, 1, 2, ..., K-1.
\end{equation}

Next, we integrate the attention weights of 12 layers, which recursively applies matrix multiplication to the original attention weights in 12 layers to obtain the final attention weights:

\begin{equation}
    A_{final} = \prod_{l=1}^{L-1} A_{l},
\end{equation}
where $L$ denotes the number of Transformer blocks in the discriminative token extractor.

Compared to the original attention weights $A_{12}$ of a single Transformer block, $A_{final}$ contains the propagation of micro-expression information from the input to the higher-level embedding, which is a better design for sense local discriminative regions. Then we select the indexes $A^1_f, A^2_f, ..., A^K_f$ relative to the maximum values of the $K$ different attention heads in $A_{final}$, and extract the corresponding indexed tokens from $Y_{12}$. Finally, the selected tokens are spliced with the class tokens to be used as the input representation of the last Transformer block:

\begin{equation}
    Y_{select}=[Y^0_l; Y^{A^1_f}_l; Y^{A^2_f}_l; ...; Y^{A^K_f}_l].
\end{equation}

Next, the last Transformer block performs the same calculations on $Y_{select} \in R^{(K+1) \times d}$ as in Eq.~\ref{eq:7} and Eq.~\ref{eq:8} to obtain $Y_{final} \in R^{(K+1) \times d}$. Finally, the MLP Head implemented by a fully connected layer and Softmax process $Y_{final}$ to obtain the prediction:

\begin{equation}
    \hat{y}= \arg\max_{i \in I} \mathrm{Softmax}(\mathrm{FC}(Y_{final}))_i,
\end{equation}
where $i$ is the index of max value in the prediction vector, and $I = (0, 1, ..., N_{class}-1)$, $N_{class}$ is the number of micro-expression classes.

\subsection{Loss Function}

Since the feature differences between micro-expression classes are minor, using only simple cross-entropy loss cannot adequately guide discriminative representation learning. Followed \cite{he2022transfg}, the contrastive loss $\mathcal{L}_{con}$ is added in addition to the cross-entropy loss, which is used to minimize the similarity of the class tokens corresponding to different labels and maximize the similarity of class tokens of samples with the same class. To prevent the loss from being dominated by simple negative samples, a constant margin $\alpha$ is introduced in the contrastive loss $\mathcal{L}_{con}$, and only pairs of negative samples with similarity more significant than $\alpha$ contribute to $\mathcal{L}_{con}$. $\mathcal{L}_{con}$ for a sample with batch size $B$ can be expressed as:

\begin{multline}
    \mathcal{L}_{con} = \frac{1}{B^2} \sum_{i}^{B} \left[ \sum_{\substack{j:  y_i=y_j}}^{B} \left(1 - \text{Sim}(Y_i, Y_j) \right) + \right. \left. \sum_{\substack{j:  y_i \ne y_j}}^{B} \left( \max \left( \text{Sim}(Y_i, Y_j) - \alpha, 0 \right) \right) \right],
\end{multline}
where $Y_i$ and $Y_j$ are pre-processed by $L2$ normalization and $\mathrm{Sim}(Y_i, Y_j)$ is the dot product of $Y_i$ and $Y_j$. The total loss can be expressed as:

\begin{equation}
    \mathcal{L} = \mathcal{L}_{ce}(y, \hat{y}) + \mathcal{L}_{con}(Y_{final}).
\end{equation}

\section{Experiments}

\subsection{Datasets}
To verify the effectiveness of our proposed method, we conducts experiments based on the CASME II, SAMM, SMIC and CAS(ME)$^3$ datasets, respectively. The detailed information of the four datasets is as follows: 

CASME II \cite{yan2014casme} is a comprehensive spontaneous MEs dataset containing 247 MEs samples collected from 26 participants. It is labeled with five emotion categories: \textit{happiness}, \textit{disgust}, \textit{surprise}, \textit{repression}, and \textit{others}, which will be used in the SDE protocol. In the CDE protocol, the samples are reclassified into three labels: \textit{positive}, \textit{negative}, and \textit{surprise}, to align with the SMIC dataset.

SAMM \cite{davison2016samm} is a spontaneous dataset comprising 159 MEs samples from 32 participants. It is labeled with 8 emotion categories: \textit{happiness}, \textit{anger}, \textit{surprise}, \textit{disgust}, \textit{fear}, \textit{sadness}, \textit{contempt}, and \textit{others}. Followed the rules of previous work, labels with few samples do not participate in training and evaluation, so five labels are left in SAMM: \textit{happiness}, \textit{anger}, \textit{surprise}, \textit{contempt}, \textit{others} for the SDE protocol. In the CDE protocol, the samples are reclassified into three labels: \textit{positive}, \textit{negative}, and \textit{surprise}, to align with the SMIC dataset.

SMIC \cite{li2013spontaneous} is a spontaneous dataset comprising 164 MEs samples from 16 participants. The dataset is labeled with 3 emotion categories: \textit{positive}, \textit{negative}, and \textit{surprise} for both CDE and SDE protocols.

CAS(ME)$^3$ \cite{li2022cas} provides about 80 hours of video, over 8 million frames, containing 1109 manually annotated MEs. The larger sample size allowed efficient validation of the MER method while avoiding database bias. It is labeled with 7 emotion categories: \textit{happiness}, \textit{anger}, \textit{surprise}, \textit{disgust}, \textit{fear}, \textit{sadness}, and \textit{others}. Followed \cite{wang2024htnet}, We reclassify these samples into three categories:  \textit{positive}, \textit{negative}, and \textit{surprise} for the SDE protocol.

Followed \cite{see2019megc}, the Single Database Experiment (SDE) and the Composite Database Evaluation (CDE) are conducted as the popular micro-expression recognition evaluation. The CDE merges CASME II, SAMM, SMIC datasets and relabels micro-expression samples in CASME II and SAMM datasets to the same with the samples in SMIC dataset. In addition, according to previous work, all evaluation is performed under the Leave-One-Subject-Out (LOSO) cross-validation evaluation protocol.

\subsection{Evaluation Metrics}

\textbf{Unweighted F1-score (UF1)}: Calculate all the true positives $TP_c$, false positives $FP_c$ and false negatives $FN_c$ of class $c$ (total $C$ classes) on the n-fold LOSO validation method, respectively, and get $F1_c$. We finally reach the UF1 by averaging the $F1_c$ of class $c$, with the following equations:

\begin{equation}
F1_c=\frac{2TP_c}{2TP_c+FP_c+FN_c}  \label{equ18}
\end{equation}
\begin{equation}
UF1=\frac{1}{C}\sum_{c}F1_c  \label{equ19}
\end{equation}

\textbf{Unweighted Average Recall (UAR)}: First calculate the accuracy $Acc_c$ for class $c$ (total $C$ classes) and the average all the $Acc_c$ to get the final UAR, the formulas are as follows:

\begin{equation}
Acc_c=\frac{TP_c}{N_c}  \label{equ20}
\end{equation}
\begin{equation}
UAR=\frac{1}{C}\sum_{c}Acc_c  \label{equ21}
\end{equation}

\subsection{Implementation details}

All experiments were conducted using PyTorch 2.0.0 on Ubuntu, the training and evaluation were done using the single NVIDIA RTX 4090 GPU (24G). On each dataset, the epoch is set to 300, the learning rate is \textit{5e-5}, and the batch size is 256, using Adam to perform parameter optimization.

\subsection{Comparison to State-of-the-art Methods}

\begin{table}[t]
    \centering
\caption{Comparative experiments with existing methods for micro-expression recognition under CDE protocol.}
\resizebox{\textwidth}{!}{
\begin{tabular}{cccccccccccc}
\toprule
\multirow{2}{*}{ Method} & \multirow{2}{*}{Params} & \multirow{2}{*}{FLOPs} & \multicolumn{2}{c}{ Full } & \multicolumn{2}{c}{ SAMM } & \multicolumn{2}{c}{ CASME II } & \multicolumn{2}{c}{ SMIC } \\
\cmidrule{4-11} &  &  & $\mathrm{UF} 1$ & UAR & $\mathrm{UF} 1$ & UAR & $\mathrm{UF} 1$ & $\mathrm{UAR}$ & $\mathrm{UF} 1$ & $\mathrm{UAR}$ \\ \midrule
\rowcolor{grey}\multicolumn{11}{l}{\textit{{Traditional micro-expression recognition methods}}} \\
LBP-TOP \cite{wang2014lbp}          & - & - & 0.5882 & 0.5785 & 0.3954 & 0.4102 & 0.7026 & 0.7429 & 0.2000 & 0.5280 \\
Bi-WOOF \cite{liong2018less}          & - & - & 0.6296 & 0.6227 & 0.5211 & 0.5139 & 0.7805 & 0.8026 & 0.5727 & 0.5829 \\
\midrule
\rowcolor{grey}\multicolumn{11}{l}{\textit{{Deep learning-based micro-expression recognition methods}}} \\
OFF-ApexNet \cite{gan2019off}       & - & - & 0.7196 & 0.7069 & 0.5409 & 0.5392 & 0.8764 & 0.8681 & 0.6817 & 0.6695 \\
STSTNet \cite{liong2019shallow}     & - & - & 0.7353 & 0.7605 & 0.6588 & 0.6810 & 0.8382 & 0.8686 & 0.6801 & 0.7013 \\
CapsuleNet \cite{van2019capsulenet} & - & - & 0.6520 & 0.6506 & 0.6209 & 0.5989 & 0.7068 & 0.7018 & 0.5820 & 0.5877 \\
Dual-Inception \cite{zhou2019dual}  & - & - & 0.7322 & 0.7278 & 0.5868 & 0.5663 & 0.8621 & 0.8560 & 0.6645 & 0.6726 \\
EMR  \cite{liu2019neural}           & - & - & 0.7885 & 0.7824 & 0.7754 & 0.7152 & 0.8293 & 0.8209 & 0.7461 & 0.7530 \\
RCN-A \cite{xia2020revealing}       & - & - & 0.7430 & 0.7190 & 0.7600 & 0.6720 & 0.8510 & 0.8120 & 0.6330 & 0.6440 \\
GEME \cite{nie2021geme}             & - & - & 0.7400 & 0.7500 & 0.6870 & 0.6540 & 0.8400 & 0.8510 & 0.6290 & 0.6570 \\
MERSiamC3D \cite{zhao2021two}       & - & - & 0.8070 & 0.7900 & 0.7480 & 0.7280 & 0.8820 & 0.8760 & 0.7360 & 0.7600 \\
FeatRef \cite{zhou2022feature}      & - & - & 0.7838 & 0.7832 & 0.7372 & 0.7155 & 0.8915 & 0.8873 & 0.7011 & 0.7083 \\
FRL-DGT \cite{zhai2023feature}      & - & - & 0.812  & 0.811  & 0.772  & 0.758  & 0.919  & 0.903  & 0.743  & 0.749 \\
GLEFFN \cite{guo2023gleffn}         & - & - & 0.8121  & 0.8208  & 0.7458  & 0.7843  & 0.8825  & 0.9110  & 0.7714  & 0.7856 \\
HTNet \cite{wang2024htnet}          & 438.51 & 214.6 & \underline{0.8603} & 0.8475 & \underline{0.8131} & 0.8124 & \underline{0.9532} & \underline{0.9516} & \underline{0.8049} & \underline{0.7905} \\
MFDAN \cite{cai2024mfdan}          & - & - & 0.8453 & \underline{0.8688} & 0.7871 & \underline{0.8196} & 0.9134 & 0.9326 & 0.6815 & 0.7043 \\
OFVIG-Net \cite{zhang2025micro}     & - & - & 0.6720 & 0.6632 & 0.6066 & 0.5787 & 0.7129 & 0.7195 & 0.6435 & 0.6400 \\
\midrule
 \textbf{Ours}       & \textbf{7.03} & \textbf{84.03} & \textbf{0.8852} & \textbf{0.8944} & \textbf{0.8383} & \textbf{0.8383} & \textbf{0.9882} & \textbf{0.9896} & \textbf{0.8279} & \textbf{0.8363} \\
\bottomrule
\end{tabular}
}
\label{tab: cecom}
\end{table}

\textbf{Experimental Results under CDE Protocol:} To validate the performance of EPIR, we conduct comparative experiments to compare EPIR with state-of-the-art methods under the CDE protocol. Furthermore, To verify the lightweight of EPIR, we calculated the number of model parameters and FLOPs for EPIR and other micro-expression recognition methods (we only considered methods released in the past three years that have public open-source implementations), the units are in millions (M). The experimental results are shown in Table~\ref{tab: cecom} (SAMM, CASME II, SMIC). In Table~\ref{tab: cecom}, - means that the results are not available in the corresponding literature. \textbf{Full} denotes the experimental results of composite training and evaluation on SAMM, CASME II, and SMIC datasets. Data in bold denotes the best result, data with an underline denotes the second-best result, and the units of Params and FLOPs are both millions (M).

As can be seen from Table \ref{tab: cecom}, firstly, compared with traditional micro-expression recognition methods, deep learning-based methods have significant performance improvements. At the same time, EPIR generally achieves better results than the existing deep learning-based methods on the three micro-expression recognition datasets, with the highest improvement in UF1 and UAR of 2.49\% and 2.56\% on the composite evaluation (Full), 2.52\% and 1.87\% on the SAMM dataset, 3.5\% and 3.8\% on the CASME II dataset, and 2.3\% and 4.58\% on the SMIC dataset. Notably, EPIR achieved improvements in most of the metrics on the three datasets using only 1.6\% of HTNet \cite{wang2024htnet} parameters. For FLOPs, EPIR has a lower time complexity while achieving superior results. Thus, we suppose that even with a very small training data scale, EPIR can achieve efficient micro-expression recognition with extremely low time and space complexity.

\begin{figure*}[t]
\begin{center}
   \includegraphics[width=\textwidth]{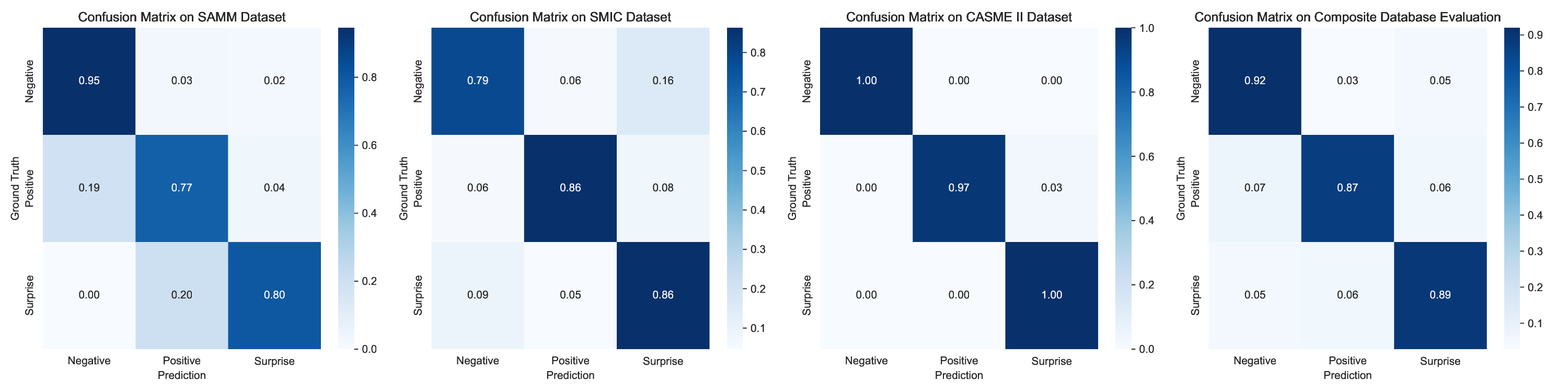}
\end{center}
   \caption{Confusion matrices for the proposed EPIR on the composite database (SAMM, SMIC, CASME II). The data in each cell denotes the accuracy for the particular class.}
\label{fig:cmcom}
\end{figure*}

Fig.~\ref{fig:cmcom} presents the confusion matrix of EPIR under the CDE protocol. In the CASME II dataset, EPIR attains a 100\% recognition accuracy for the Negative and Surprise classes.

\begin{figure}[t]
\begin{center}
   \includegraphics[width=0.4\textwidth]{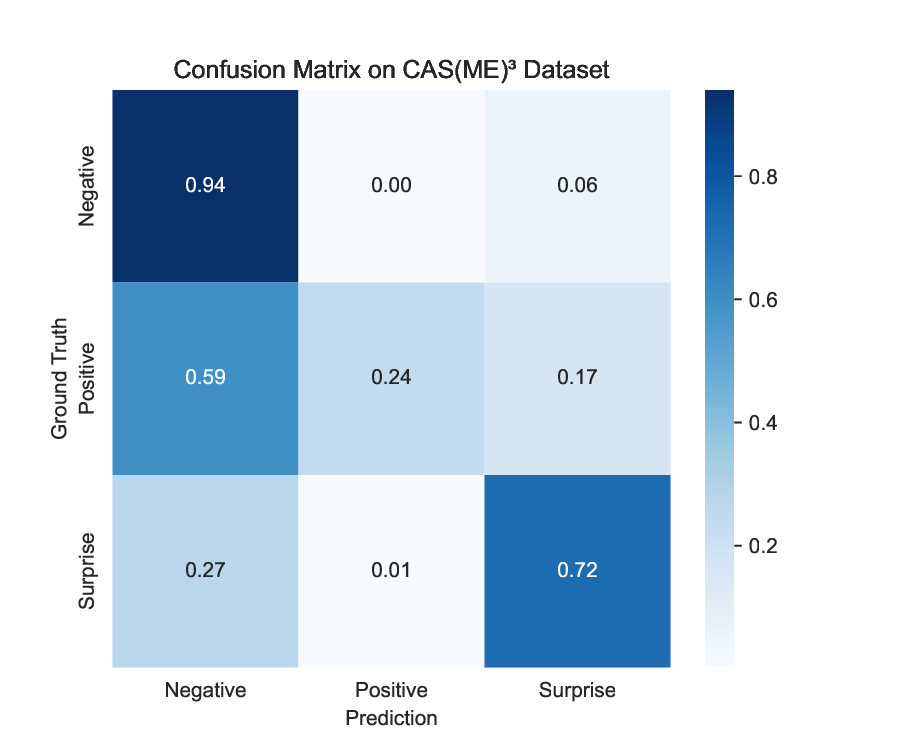}
\end{center}
   \caption{Confusion matrices for the proposed EPIR on the CAS(ME)$^3$. The data in each cell denotes the accuracy for the particular class.}
\label{fig:cm3}
\end{figure}

\begin{table}[t]
    \centering
\caption{Comparative experiments with existing methods for micro-expression recognition on CAS(ME)$^3$ under SDE protocol.}
\begin{tabular}{ccccc}
\toprule
\multirow{2}{*}{ Method} & \multirow{2}{*}{Params} & \multirow{2}{*}{FLOPs} & \multicolumn{2}{c}{ CAS(ME)$^3$} \\
\cmidrule{4-5} &  &  & $\mathrm{UF} 1$ & UAR \\ \midrule
STSTNet \cite{liong2019shallow}     & - & - & 0.3795 & 0.3792 \\
FeatRef \cite{zhou2022feature}      & - & - & 0.3493 & 0.3413 \\
$\mu$-BERT \cite{nguyen2023micron}  & 333.4 & 414.5 & 0.5604 & \underline{0.6125} \\
HTNet \cite{wang2024htnet}          & 438.51 & 214.6 & \underline{0.5767} & 0.5415 \\
\midrule
\textbf{Ours}       & \textbf{7.03} & \textbf{84.03} & \textbf{0.6727} & \textbf{0.6345} \\
\bottomrule
\end{tabular}
\label{tab: ce3}
\end{table}

\begin{table*}[t]
    \centering
\caption{Comparative experiments for micro-expression recognition under SDE protocol.}
\resizebox{\textwidth}{!}{
\begin{tabular}{ccccccccc}
\toprule
\multirow{2}{*}{ Method} & \multirow{2}{*}{Params} & \multirow{2}{*}{FLOPs} & \multicolumn{2}{c}{ SAMM } & \multicolumn{2}{c}{ CASME II } & \multicolumn{2}{c}{ SMIC } \\
\cmidrule{4-9} &  &  & $\mathrm{UF} 1$ & UAR & $\mathrm{UF} 1$ & $\mathrm{UAR}$ & $\mathrm{UF} 1$ & $\mathrm{UAR}$ \\ \midrule
\rowcolor{grey}\multicolumn{9}{l}{\textit{{Traditional micro-expression recognition methods}}} \\
LBP-TOP \cite{wang2014lbp}          & - & -  & 0.3589 & 0.3968 & 0.3589 & 0.3968 & 0.3421 & 0.4338 \\
MDMO \cite{liu2015main}             & - & -  & - & - & 0.4966 & 0.5169 & 0.5845 & 0.5897 \\
Bi-WOOF \cite{liong2018less}        & - & -  & - & - & 0.6100 & 0.5885 & 0.6200 & 0.6220 \\
\midrule
\rowcolor{grey}\multicolumn{9}{l}{\textit{{Deep learning-based micro-expression recognition methods}}} \\
Graph-TCN \cite{lei2020novel}       & - & -  & 0.6985 & 0.7500 & 0.7246 & 0.7398 & - & - \\
LGCcon \cite{li2020joint}     & - & -  & 0.3400 & 0.4090 & 0.6400 & 0.6502 & - & - \\
AUGCN+AUFusion \cite{lei2021micro} & - & -  & 0.7045 & 0.7426 & 0.7047 & 0.7427 & - & - \\
SLSTT \cite{zhang2022short}       & - & -  & 0.6400 & 0.7239 & 0.7530 & 0.7581 & 0.7240 & 0.7371 \\
I3D+MOCO \cite{wang2014lbp}      & - & -   & 0.5436  & 0.6838  & 0.7366  & 0.7630  & 0.7492  & 0.7561 \\
SRMCL \cite{bao2024boosting}      & - & -   & 0.6599  & 0.7463  & 0.8286  & 0.8320  & 0.7887  & 0.7898 \\
\midrule
\rowcolor{grey}\multicolumn{9}{l}{\textit{{Large language model-based micro-expression recognition methods}}} \\
MELLM \cite{zhang2025mellm}      & - & -   & -  & -  & 0.4849  & 0.5337  & -  & - \\

\midrule
 \textbf{Ours}       & \textbf{7.03} & \textbf{84.03} & \textbf{0.8011} & \textbf{0.8052} & \textbf{0.8712} & \textbf{0.8466} & \textbf{0.8025} & \textbf{0.8213} \\
\bottomrule
\end{tabular}
}
    \label{tab: cecomsde}
\end{table*}

\textbf{Experimental Results under SDE Protocol:} Table~\ref{tab: ce3} and Table~\ref{tab: cecomsde} show the results among EPIR with existing methods under SDE protocol, Fig.~\ref{fig:cm3} illustrates the confusion matrix on the CAS(ME)$^3$ dataset. In Table~\ref{tab: ce3} and Table~\ref{tab: cecomsde}, - means that the results are not available in the correspond literature. Data in bold indicates the best-performing result, data with an underline indicates the second-best result, and the units of Params and FLOPs are both millions (M). Firstly, compared with traditional methods, deep learning-based methods can still be significantly ahead. Compared with the micro-expression recognition method based on large language models, conventional models are obviously more suitable for micro-expression recognition at present. Micro-expression recognition is a fine-grained task, and the current large language model's cognition is not sufficient to complete it. At the same time, due to the large scale, more resources will be consumed. Among the methods based on deep learning, EPIR can show better recognition results than the SOTA methods on the four datasets under SDE protocol, with the improvement in UF1 and UAR of 9.66\% and 5.52\% on the SAMM dataset, 4.26\% and 1.46\% on the CASME II dataset, 1.38\% and 3.15\% on the SMIC dataset, and 9.6\% and 2.2\% on the CAS(ME)$^3$ dataset. For efficiency, EPIR achieved improvements on the CAS(ME)$^3$ datasets using only 2.1\% of $\mu$-Bert \cite{nguyen2023micron}.

\subsection{Ablation Study}

\begin{table*}[t]
    \centering
\caption{Ablation experiments for micro-expression recognition.}
\resizebox{\textwidth}{!}{
\begin{tabular}{ccccccccc}
\toprule
\multirow{2}{*}{ Method} & \multicolumn{2}{c}{ Full } & \multicolumn{2}{c}{ SAMM } & \multicolumn{2}{c}{ CASME II } & \multicolumn{2}{c}{ SMIC } \\
\cmidrule{2-9} & $\mathrm{UF} 1$ & UAR & $\mathrm{UF} 1$ & UAR & $\mathrm{UF} 1$ & $\mathrm{UAR}$ & $\mathrm{UF} 1$ & $\mathrm{UAR}$ \\ \midrule
 Baseline       & 0.7325 & 0.7214 & 0.6318 & 0.6307 & 0.8511 & 0.8532 & 0.6532 & 0.6767 \\
 $+$DTSM     & 0.7736 & 0.7838 & 0.7351 & 0.7253 & 0.8741 & 0.8703 & 0.7147  & 0.7052 \\
 $+$ITALS         & 0.8021 & 0.8165 & 0.7962 & 0.7815 & 0.8753 & 0.8712 & 0.7501 & 0.7326 \\
 $+$SPT         & 0.8513 & 0.8627 & 0.8135 & 0.7831 & 0.9797 & 0.9714 & 0.8031 & 0.8122 \\
 $+$DNSPT         & 0.8759 & 0.8747 & 0.8288 & 0.7941 & 0.9808 & 0.9791 & 0.8167 & 0.8215 \\
 $+$Token Integration         & \textbf{0.8852} & \textbf{0.8944} & \textbf{0.8303} & \textbf{0.8303} & \textbf{0.9882} & \textbf{0.9896} & \textbf{0.8279} & \textbf{0.8363} \\
\bottomrule
\end{tabular}
}
    \label{tab: aecom}
\end{table*}

\begin{table}[t]
    \centering
\caption{Ablation experiments for micro-expression recognition on CAS(ME)$^3$ dataset.}
\begin{tabular}{ccc}
\toprule
\multirow{2}{*}{ Method } & \multicolumn{2}{c}{ CAS(ME)$^3$} \\
\cmidrule{2-3} & $\mathrm{UF} 1$ & UAR \\ \midrule
 Baseline       & 0.3647 & 0.3129 \\
 $+$DTSM     & 0.4497 & 0.4792 \\
 $+$ITALS         & 0.5318 & 0.5132 \\
 $+$SPT         & 0.6233 & 0.5613 \\
 $+$DNSPT         & 0.6604 & 0.6169 \\
 $+$Token Integration         & \textbf{0.6727} & \textbf{0.6345} \\
\bottomrule
\end{tabular}
\label{tab: ae3}
\end{table}

We conducted ablation experiments to verify the role of various components in EPIR. The quantitative results are shown in Table \ref{tab: aecom} and Table \ref{tab: ae3}. In Table \ref{tab: aecom} and Table \ref{tab: ae3}, \textbf{Full} denotes the experimental results of composite training and evaluation on SAMM, CASME II, and SMIC datasets. The specific design of the different experimental control groups is as follows: 

\begin{itemize}
    \item \textbf{Baseline} denotes using the Vision Transformer \cite{dosovitskiy2020image} only; 

    \item \textbf{+DTSM} denotes the integration of the DTSM before the last Transformer block in the discriminative token extractor, along with the $\mathcal{L}_{con}$ during training; 

    \item \textbf{+ITALS} denotes modifying the self-attention of the Transformer block in \textbf{+DTSM} group to ITALS; 

    \item \textbf{+SPT} denotes that the shifted patch tokenization (SPT) is added to the \textbf{+ITALS} group; 

    \item \textbf{+DNSPT} denotes that the dual norm projection module is added to the SPT based on the \textbf{+SPT} group.
    
    \item \textbf{+Token Integration} denotes that the token integration module is added based on the \textbf{+DNSPT} group, i.e., EPIR.
\end{itemize}

\begin{figure*}[t]
\begin{center}
   \includegraphics[width=\textwidth]{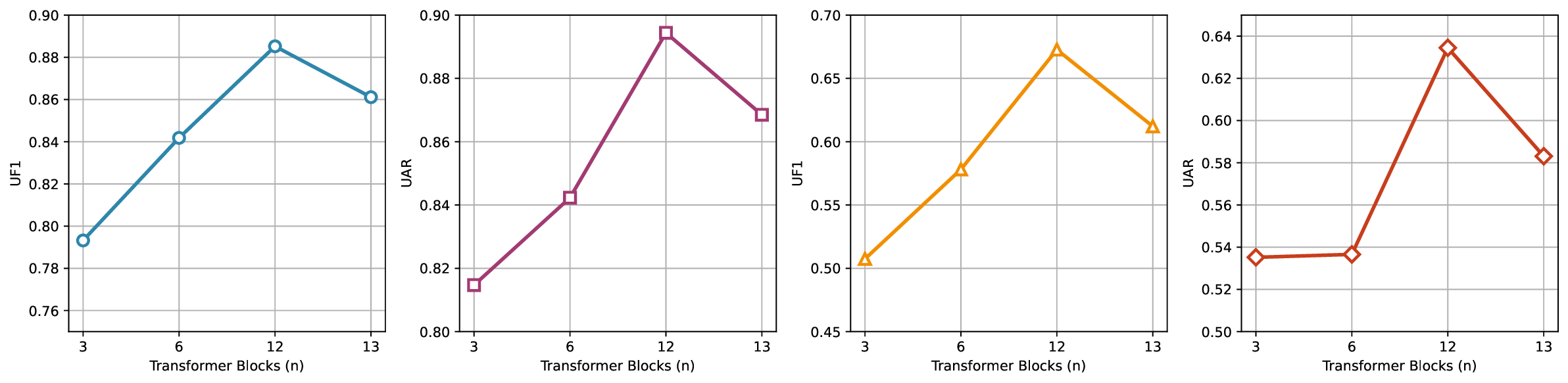}
\end{center}
   \caption{Ablation experiments on the number of Transformer blocks, the two subfigures on the left are the UAR and UF1 metrics under the CDE protocol, and the two subfigures on the right are the UAR and UF1 metrics on the CAS(ME)$^3$ dataset. The abscissa is the number n of Transformer blocks in all cases.}
\label{fig: blockae}
\end{figure*}

\begin{figure}[t]
\begin{center}
   \includegraphics[width=0.6\textwidth]{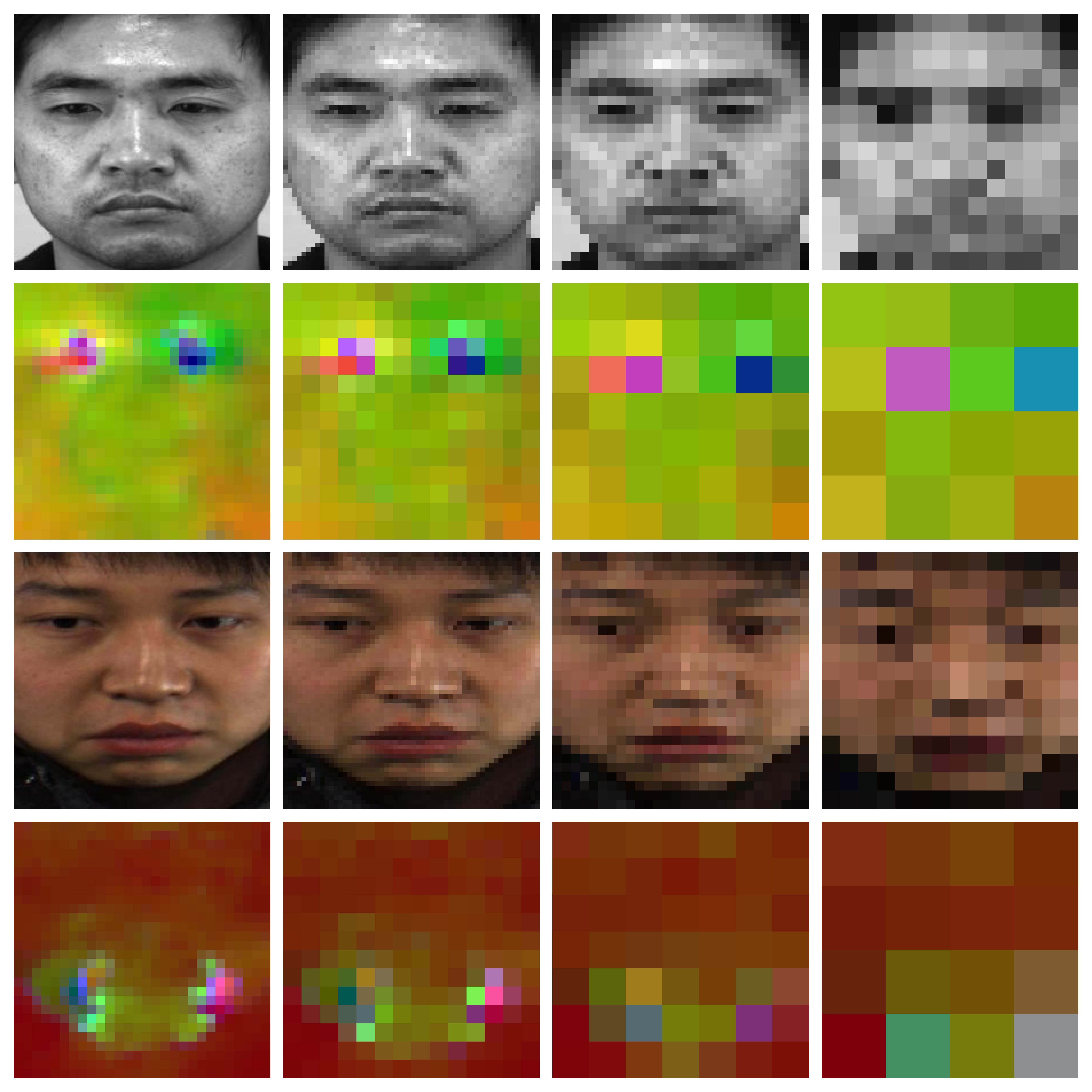}
\end{center}
   \caption{Actual effect of token integration module on micro-expression samples. We take two micro-expression samples (first and third rows) and their corresponding optical flow feature maps (second and fourth rows), from left to right, with integration rates of 0\%, 30\%, 60\%, and 80\%, respectively.}
\label{fig: tokenvisual}
\end{figure}

\textbf{The role of DTSM.} The experimental results in Table \ref{tab: aecom} and Table \ref{tab: ae3} denote that compared to the ordinary vision Transformer \cite{dosovitskiy2020image}, DTSM achieves better micro-expression recognition due to its better capability of local discriminative feature extraction. The combination of DTSM and $\mathcal{L}_{con}$ enables the model to focus on discriminative features of different micro-expression classes.

\textbf{The role of ITALS.} The experimental results between \textbf{+DTSM} group and \textbf{+ITALS} group show that by improving the self-attention to ITALS, the learnable scaling and diagonal masking successfully make softmax's output distribution more discriminative. This not only makes the model converge as quickly as possible from small-scale data but also helps the DTSM select discriminative micro-expression representations more easily.

\textbf{The role of DNSPT.} The experimental results between \textbf{+ITALS} group and \textbf{+SPT} group show that by performing spatial transformations on the input optical flow features before tokenization, SPT implicitly augments all input representations, allowing it to achieve good performance even when dealing with small-scale datasets. Additionally, the results between \textbf{+SPT} group and \textbf{+DNSPT} group show that dual normalization projection effectively enhances training stability by controlling the mean gradient norm during model training in the process of projecting optical flow patches into visual tokens, contributing significantly to the robustness and generalization of EPIR.

\textbf{The role of the Token Integration.} The experimental results between \textbf{+DNSPT} group and \textbf{+Token Integration} group show that token integration has promoted the recognition performance of the model on the four datasets. By increasing the density of micro-expression information, this module not only reduces the computational complexity of the model, but also ensures or even improves the recognition effect of the model.

In addition, we verify the actual effect of token integration module on micro-expression samples, and the experimental results are shown in Fig.~\ref{fig: tokenvisual}. We can clearly find that when the integration rate of tokens is 30\%, the information of micro-expressions is not significantly missing, and the balance between computational complexity and recognition performance can be achieved in this case. However, when the integration rate reaches 60\%, the micro-expression information has been significantly missing, although the computational complexity is reduced, the recognition performance has been greatly reduced. When the integration rate reaches 80\%, the information of micro-expression has nearly disappeared, and it cannot be represented and recognized.

\textbf{The role of the number of Transformer blocks.} To verify the role of the number of Transformer blocks on the recognition performance, we conducted ablation experiments as shown in Fig.~\ref{fig: blockae}. We find that when the number of Transformer block is less than 13, the recognition performance is proportional to the number of Transformer blocks. When the number of Transformer blocks is greater than or equal to 13, the recognition performance of the model is inversely proportional to the number of Transformer blocks, we eventually chose to set the number of Transformer blocks to 13.

\section{Conclusion}
In this paper, we propose EPIR for micro-expression recognition. EPIR is a new design based on PTSR. To further realize the lightweight of the model while ensuring the recognition performance, we selectively integrate tokens in Transformer blocks, and successfully achieve more efficient micro-expression recognition. We conduct comprehensive large-scale experiments to evaluate the model performance, extensive experimental results on SAMM, CASME II, SMIC, and CAS(ME)$^3$ datasets demonstrate that EPIR achieves significant performance improvement under the difficult conditions of extremely low computational complexity and small-scale datasets, demonstrating the superiority and efficiency of EPIR. We hope that this work can bring inspiration to the micro-expression recognition community.

\section{CRediT authorship contribution statement}
Junbo Wang: Writing - original draft, Writing - review \& editing, Conceptualization, Methodology, Formal analysis, Supervision, Funding acquisition, Project administration, Resources. Liangyu Fu: Writing - original draft, Conceptualization, Methodology, Software, Data curation, Investigation, Validation, Formal analysis, Visualization. Yuke Li: Writing - review \& editing, Supervision, Resources, Funding acquisition. Yining Zhu: Writing - review \& editing, Supervision, Resources, Funding acquisition. Xuecheng Wu: Writing - review \& editing, Resources. Kun Hu: Writing - review \& editing, Resources.

\section{Declaration of competing interest}
The authors declare that they have no known competing financial interests or personal relationships that could have appeared to influence the work reported in this paper.

\section{Data availability}
Data will be made available on request.

\section{Declaration of generative AI and AI-assisted technologies in the manuscript preparation process}
This work did not use generative AI and AI-assisted technologies in the manuscript preparation process.

\bibliographystyle{elsarticle-num} 
\bibliography{references}

\end{document}